\documentclass[conference]{IEEEtran}
\usepackage{comment}
\IEEEoverridecommandlockouts
\usepackage{cite}
\usepackage{amsmath,amssymb,amsfonts}
\usepackage{algorithmic}
\usepackage{graphicx}
\usepackage{textcomp}
\usepackage{xcolor}
\def\BibTeX{{\rm B\kern-.05em{\sc i\kern-.025em b}\kern-.08em
    T\kern-.1667em\lower.7ex\hbox{E}\kern-.125emX}}
\begin{document}

\title{Optical Text Recognition in Nepali and Bengali: A Transformer-based Approach\\}

%\begin{comment}
\author{\IEEEauthorblockN{S M Rakib Hasan}
\IEEEauthorblockA{\textit{Department of Computer Science and Engineering} \\
\textit{BRAC University}\\
Dhaka, Bangladesh \\
sm.rakib.hasan@g.bracu.ac.bd}
\and
\IEEEauthorblockN{Aakar Dhakal}
\IEEEauthorblockA{\textit{Department of Computer Science and Engineering} \\
\textit{BRAC University}\\
Dhaka, Bangladesh \\
aakar.dhakal@g.bracu.ac.bd}

\and
\IEEEauthorblockN{Md Humaion Kabir Mehedi} \IEEEauthorblockA{\textit{Department of Computer Science and Engineering} \\
\textit{BRAC University}\\
Dhaka, Bangladesh \\
humaion.kabir.mehedi@g.bracu.ac.bd}
\and
\IEEEauthorblockN{Annajiat Alim Rasel}
\IEEEauthorblockA{\textit{Department of Computer Science and Engineering} \\
\textit{BRAC University}\\
Dhaka, Bangladesh \\
annajiat@gmail.com}}
%\end{comment}
\IEEEoverridecommandlockouts
\IEEEpubid{\makebox[\columnwidth]{979-8-3503-0544-9/23/\$31.00 ©2023 IEEE \hfill}
\hspace{\columnsep}\makebox[\columnwidth]{ }}
\maketitle
\IEEEpubidadjcol
\begin{abstract}
Efforts on the research and development of OCR systems for Low-Resource Languages are relatively new. Low-resource languages have little training data available for training Machine Translation systems or other systems. Even though a vast amount of text has been digitized and made available on the internet the text is still in PDF and Image format, which are not instantly accessible. This paper discusses text recognition for two scripts: Bengali and Nepali; there are about 300 and 40 million Bengali and Nepali speakers respectively. In this study, using encoder-decoder transformers, a model was developed, and its efficacy was assessed using a collection of optical text images, both handwritten and printed. The results signify that the suggested technique corresponds with current approaches and achieves high precision in recognizing text in Bengali and Nepali. This study can pave the way for the advanced and accessible study of linguistics in South East Asia.
\end{abstract}


\section{Introduction}
With the advent of ever-changing digital media and technology in recent years, there has been a sharp rise in the demand for automated text detection. Optical Character Recognition(OCR) is a technology that enables the recognition of printed or handwritten text characters from scanned images. It improves access to information, preserves rare texts and enables the development of language technologies such as voice assistants and machine translation systems. Despite numerous works on text recognition in various languages, Bengali and Nepali text recognition has not received enough attention due to its resource deficiency or being morphologically complex. Bengali and Nepali are two extensively spoken languages in South Asia, and understanding them is essential for computer translation, document digitisation, and language processing. In this paper, an image-based method is suggested for identifying Bengali and Nepali written text. A model is developed using Microsoft TrOCR, an encoder-decoder based transformer architecture as the base model and its efficacy was assessed using a collection of text images. The outcomes demonstrate that the suggested technique achieves high precision in deciphering typed text in Bengali and Nepali. This study could aid in the advancement of linguistic technology in South Asia and be useful in a number of industries, including automation, administration, and education. This paper discusses the previous research, some unique characteristics of Bengali and Nepali text, segmentation and feature extraction methods followed by the experimental results and conclusion.

\section{Previous Works}

There have been many OCR models for low-resource languages like Bengali or Nepali and some remarkable research exists in this field. An end-to-end word identification system for handwritten Bengali words from pictures is introduced in the paper \cite{b1}. Deep convolutional neural networks (CNNs) are used by authors as feature extractors, followed by RNNs and a completely connected layer that produces the end prediction. The Connectionist Temporal Classification (CTC) loss function is used to teach the algorithm. The efficacy of four distinct baseline models—Xception, NASNet, MobileNet, and DenseNet—as feature extractors are examined in this article. The writers come to the conclusion that for Bengali handwritten OCR, deeper systems with residuals work better. The BanglaWriting dataset, a reputable Bengali dataset, is used to assess the suggested technique. With a word recognition accuracy of 90.3\%, the authors describe encouraging findings. 

The study \cite{b2} presents an optical character recognition (OCR) framework for printed Bengali and English text. The system was developed utilizing a solitary, 128-unit hidden bidirectional long short-term memory (BLSTM) architecture with connectionist temporal classification (CTC). The proposed methodology consists of a dual-phase process. In the initial stage, the vertical strip-based projection profile valleys of the document components were regarded as the primary suppositions for the line end/beginning. The aforementioned hypothesis underwent further development in the subsequent stage, wherein depictions of objects and additional artefacts were excluded. The text lines are subjected to normalization to achieve a height of 48 pixels before being presented to the BLSTM-CTC-based OCR system for training and validation. Performance evaluation employs test data that serves as a reference standard. During the testing phase, the CTC functions as the classifier and produces the most probable classifications for a given input sequence as the final outcome. The OCR system's accuracy at the character level was 99.32\%, while its accuracy at the word level was 96.65\%.

The difficulties of optical character recognition (OCR) for Bangla text are discussed in the work\cite{b3}, along with a useful character segmentation method. Using vertical and curved scanning, the segmentation approach uses line, word, and character segmentation. The segmented character extraction procedure utilizing the flood-fill method is also covered in the article. The suggested method had 99\% character segmentation accuracy, 99.8\% line segmentation accuracy, and 99.5\% word segmentation accuracy. In the publication, character segmentation accuracy experimental findings are presented. For five separate pictures, 99\% accuracy was attained. The suggested method can enhance the accuracy of OCR for printed Bangla text. 

A low-cost convolutional neural network design for Bengali handwritten character identification is suggested by the writers in the study \cite{b4}. For the training phase, the researchers used openly accessible standard datasets, such as the CMATERdb Bengali handwritten character dataset, and created various dataset forms in accordance with prior research. BengaliNet, the suggested design, outperformed earlier work by a significant margin, achieving a total accuracy of 96-99\% for various databases of Bengali characters. The study aids in the creation of an instrument that can automatically and effectively recognize Bengali handwriting symbols. 

The paper\cite{b5} suggests a customized version of a cutting-edge deep learning methodology to identify handwritten Bengali symbols. The suggested approach makes use of transfer learning on the ImageNet dataset along with a pre-trained Resnet-50 deep convolutional neural network model. By altering the input image sizes, the technique also changes the one-cycle strategy to guarantee quicker training. The 84-class BanglaLekha-Isolated dataset is used to assess the suggested methodology. The findings demonstrate that the suggested approach outperforms other contemporary methods, achieving 97.12\% accuracy in just 47 epochs. The architecture and regularizations, including batch normalization, dropout, learning rate schedule, and optimizers, used to boost speed are also covered in the article. The authors come to the conclusion that ResNet-50 is successful at classifying Bengali handwritten characters using their suggested technique. 

A novel feature set for handwritten Bangla alphabet detection is described in the paper\cite{b6}. The feature collection comprises 132 features, including quad-tree-based longest run features, distance-based features, modified shadow features, octant and centroid features, and features based on centroid and distance. The features are calculated from binary pictures of Bangla alphabetic letters with a dimension of 64x64 pixels. Four longest-run features are calculated for every sub-image at every point of the quad-tree structure. The novelty of the present study is the partitioning of any character sequence using the CG-based quad-tree structure. From the 75.05\% seen in the authors' earlier work to 85.40\% on 50 character classes with an MLP-based classifier on the same dataset, the identification performance using this feature set rises significantly.

The authors of the study \cite{b7} established a novel benchmark for 60 low-resource languages that possess low-resource scripts. This was accomplished through the utilization of authentic and synthetic data that was augmented with noise. The researchers conducted a benchmarking study to identify the most common errors in Optical Character Recognition (OCR) models that are utilized for both academic research and commercial applications. This study utilized state-of-the-art OCR mistakes to evaluate the impact of fine-tuning machine translation (MT) models with OCR-ed data in comparison to pre-trained MT models and MT models fine-tuned with initial data. The study aimed to contrast the MT models and determine the effect of SOTA OCR mistakes on them. The principal finding is that the utilization of back translation in optical character recognition-processed monolingual data results in an improvement in machine translation. Overall, the majority of current optical character recognition (OCR) models exhibit a high level of recognition accuracy, which allows for the training of machine translation (MT) models. However, there are certain scripts, such as Perso Arabic, that present challenges to OCR models. Additionally, while this augmentation process is generally robust to various types of errors, it may be susceptible to substitutions in particular. This work will enable future research endeavours focused on data augmentation for machine translation utilizing OCR documents.

The study\cite{b14} introduces a novel Optical Character Recognition (OCR) system designed for scripted Bangla characters. The system has been constructed utilizing a Convolutional Neural Network (CNN) for the purpose of character recognition. The system also integrates various techniques such as binarization and noise reduction to enhance the precision of the results. The results of the performance evaluation indicate that the proposed system exhibits superior accuracy in comparison to existing methods with respect to line detection, word segmentation, and character recognition. The study involved conducting an experiment on the recognition of a maximum number of distinct characters, which was 202. Additionally, the number of training and test images utilized in the experiment exceeded those used in prior research. A precision rate of 99.33\% was attained. The system exhibits robust performance in diverse conditions, including images that are affected by noise or blurriness. The duration of computation may pose a challenge contingent upon the capacity of the system in which it is executed, and the system's efficacy may be compromised in instances where the image is excessively rotated.

\section{Characteristics of Bengali and Nepali Texts}

By nature, Bengali, Nepali and other oriental languages have quite irregular and unique forms than Latinized languages. These features make the feature extraction and mapping of these languages difficult. However, in its written form, the Bengali language has a total of 50 letters in its alphabet, 11 being vowels and 39 consonants and the Nepali language has the same number of vowels with 33 consonants. Both these languages have some similarities in their written form which makes their interpretation harder, described as follows:
\subsection{Dependant Letters}
There are vowels and consonants joined with the consonants to produce the sounds in the words. So, the regular shape of the letters gets changed very frequently. These can be in many different shapes(each different for short and long sounds), which can get very confusing for the machine to interpret. These also change the normal distancing between letters which makes it more difficult to segment the letters. Fig-1 and Fig-2 depict these patterns for both languages
\begin{figure}[h!]
\centerline{\includegraphics{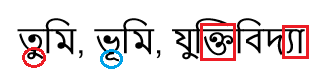}}
\caption{Dependant letters in Bengali Text.}
\label{fig}
\end{figure}
\begin{figure}[h!]
\centerline{\includegraphics{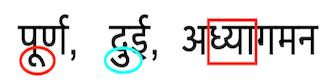}}
\caption{Dependant letters in Nepali Text.}
\label{fig}
\end{figure}

\subsection{Overlapping Characters}

Often the letters of these alphabets do not remain bounded to a certain region and enter the region of its next character. It happens because the dependent letters join and get into the next letters. It is shown in figure-3 and figure-4:

\begin{figure}[h!]
\centerline{\includegraphics{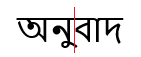}}
\caption{Overlapping Character in a Bengali word.}
\label{fig}
\end{figure}

\begin{figure}[h!]
\centerline{\includegraphics{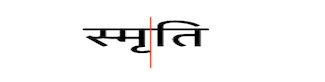}}
\caption{Overlapping Character in a Nepali word.}
\label{fig}
\end{figure}

\subsection{Word Separation}
There is a line on the letters of each word which keeps the word together. This line, known as the “Matra” line can determine the words separately. However, there are letters which skip the line above them, making the feature extraction difficult. Also, above the matra line, there may be parts of a letter and below there may be vowels. The figures below show this phenomenon.

\begin{figure}[h!]
\centerline{\includegraphics{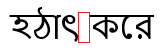}}
\caption{Separation of words in Bengali texts.}
\label{fig}
\end{figure}

\begin{figure}[h!]
\centerline{\includegraphics{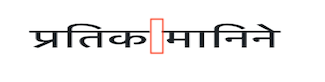}}
\caption{Separation of words in Nepali texts.}
\label{fig}
\end{figure}

\subsection{Confusing Shapes}
Both languages have some letters that are often mistaken to be the same. For a machine, it becomes tough to differentiate between these letters having very low data to train on. The following figures show these letters in both languages.

\begin{figure}[h!]
\centerline{\includegraphics{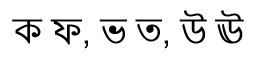}}
\caption{Some confusing letters in Bengali.}
\label{fig}
\end{figure}

\begin{figure}[h!]
\centerline{\includegraphics{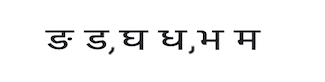}}
\caption{Some confusing letters in Nepali.}
\label{fig}
\end{figure}

\section{Methodology}
Apart from common OCR techniques in low-resourced languages that use CNNs or RNNs, a transformer-based method is proposed. The Transformer models, in contrast to the features retrieved by the CNN-like network, do not contain picture-specific inductive biases but do the image processing as a series of patches, making it simpler for the model to pay attention to either the entire image or the independent patches. In this study, the transformer model TrOCR\cite{b9} is used as the base model and extended to extract features from Bengali and Nepali images and map these features to texts. TrOCR uses typical encoder-decoder methods. It can be divided into two distinct parts

\subsection{Feature Extraction Encoder}
An image transformer is used in this part which takes some input images and extracts their features as outputs. Here, the VisionEncoderDecoder model with the pre-trained Vision Transformer (ViT)\cite{b10} is used as the encoder. The google/vit-base-patch16-384 from huggingface was used which is pre-trained on 14 million images and fine-tuned on one million images. The model receives images as a series of linearly embedded, fixed-size patches with a 16x16 resolution. In order to employ a sequence for classification tasks, a [CLS] token is also added to the beginning of the sequence. Before feeding the sequence to the layers of the Transformer encoder, one may additionally include absolute position embeddings. An instance of utilizing a dataset of labelled images involves training a conventional classifier through the stacking of a linear layer onto the pre-existing encoder. Through pre-training, the model acquires an internal representation of images that can be leveraged to extract features that are advantageous for subsequent tasks. The final hidden state of the [CLS] token is often regarded as a comprehensive representation of an entire image and is therefore commonly augmented with a linear layer. Figure 9 depicts the architectural design of the model. After the feature extraction, the language modelling is done by the decoder.

\begin{figure}[h!]
\includegraphics[width=0.5\textwidth]{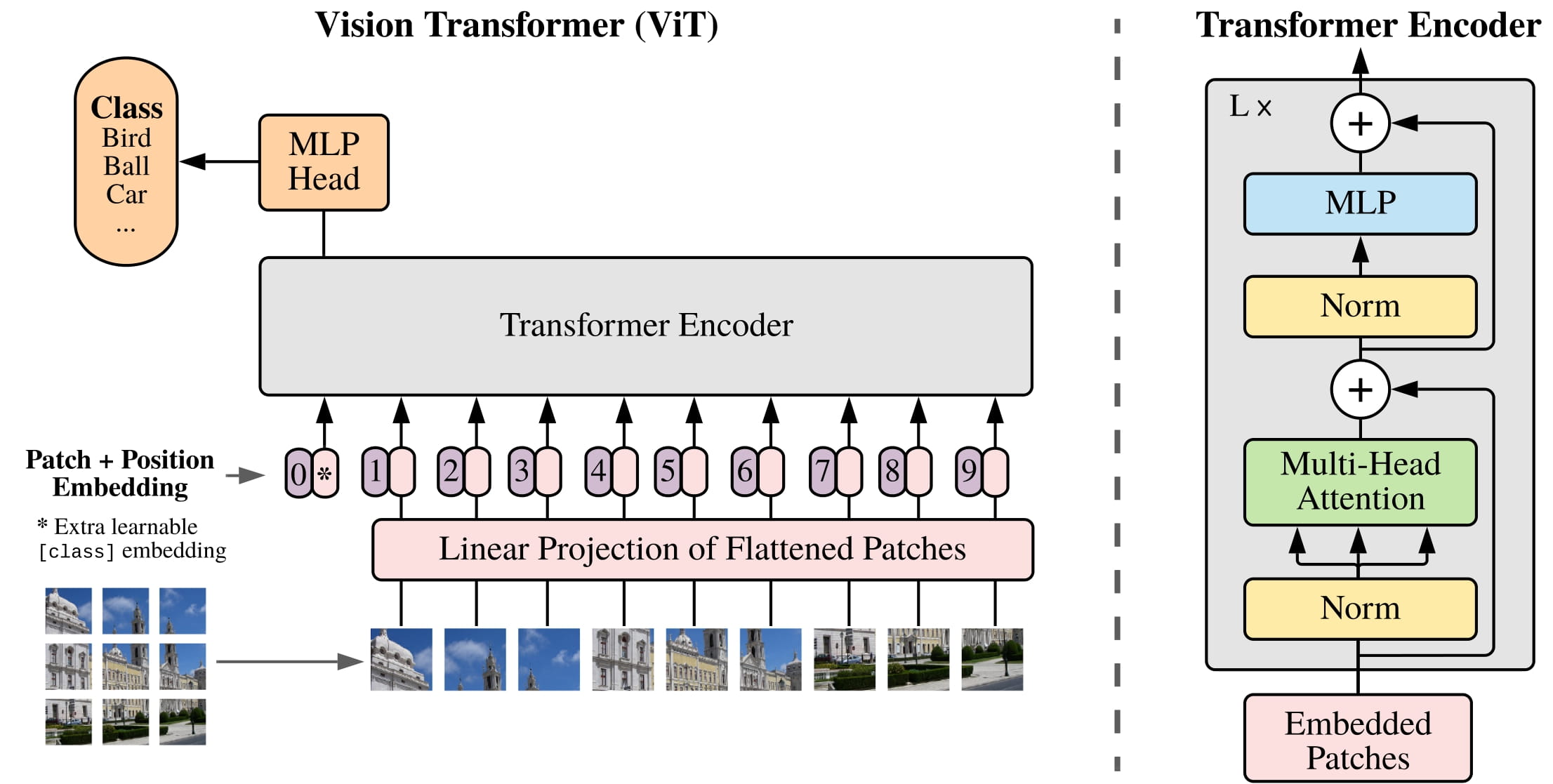}
\caption{The ViT architecture\cite{b8}.}
\label{fig}
\end{figure}

\subsection{Language Modelling}
A text transformer is used to map the extracted features to texts. In this instance, the xlm-roberta-base decoder has been chosen. This is a multilingual text transformer model that has been trained on 2.5 terabytes of data from 100 languages. There are not many language decoder models specifically for Bengali and Nepali. Therefore, this model is selected that can perform the required tasks in both languages. This is a multilingual rendition of RoBERTa. It was pretrained with the objective of Masked language modelling (MLM). Using a sentence as input, the model conceals 15\% of the words at random before running the entire sentence through the model and attempting to predict the hidden words. This differs from conventional recurrent neural networks (RNNs), which typically see the words one after the other, and autoregressive models such as GPT, which conceal the future tokens internally. It enables the model to learn a representation of the sentence that is bidirectional. Thus, the model learns an internal representation of 100 languages that can then be used to derive features useful for downstream tasks[11].
\begin{figure}[h!]
\includegraphics[width=0.5\textwidth]{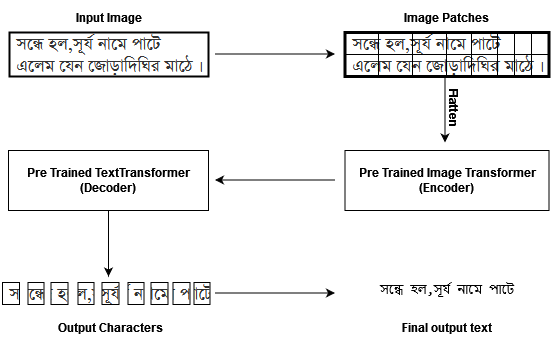}
\caption{Proposed workflow}
\label{fig}
\end{figure}
\section{Experimental Setup}
\subsection{Datasets}
For the Bengali language, the open-source BanglaWriting\cite{b12} dataset is used for fine-tuning the transformer model. It includes the single-page handwriting of 260 people of various ages and personalities This data set includes a total of 21,234 words and 32,784 characters. Additionally, this dataset comprises 5,470 unique Bangla words. The dataset also contains 261 comprehensible overwriting and 450 handwritten strikes and errors, in addition to the usual words. The word labels are generated manually. For the Nepali language, a manually prepared dataset is taken which includes 50 images from different people. It contains around 7 thousand words with about 10 thousand characters. It also has many unique words and the word labels are manually generated. Both language datasets have been used separately for training and validation.

\subsection{Data Preprocessing}
The selected dataset contained images of varying sizes. Each image was resized to 384x384 pixels for input to the encoder, ViT. The images had already been denoised, so the denoising step was omitted. Since the encoder performs effectively with regular images, the data was not converted to grayscale. The images were enhanced through random flipping and random rotation (-5 to 5 degrees).

\subsection{Training} 
The model was trained on the dataset for 2000 epochs for each language using a batch size of 4 images. Training took place on a machine equipped with a 3.6GHz CPU, 16 GB RAM, and an RTX 3070 GPU. Optimization was performed using the AdamW algorithm, a stochastic gradient descent technique that incorporates adaptive prediction of first-order and second-order moments, along with a weight decay method according to the techniques\cite{b13}. Visualization of training and validation loss was accomplished using the WandB and DataCollator libraries.
\subsection{Testing} 
The performance of the proposed trained Optical Character Recognition (OCR) model was assessed by analyzing a set of 10 images obtained from various sources. The chosen images were curated to exemplify a broad spectrum of font styles, sizes, and orientations that are frequently encountered in practical situations.
The initial step in the testing procedure involved image preprocessing, which entailed resizing the images to a uniform size of 512x512 pixels and converting them to grayscale. Subsequently, the preprocessed images were inputted into the OCR model, and the resulting recognized text was documented.
In order to assess the precision of the optical character recognition (OCR) model, the subsequent metrics were computed:
\begin{itemize}
	\item The metric of character-level accuracy evaluates the proportion of accurately identified characters within the completely recognized text. The computation involves the division of the count of accurately identified characters by the overall count of characters present in the reference text.

	\item The metric of word-level accuracy evaluates the proportion of accurately identified words within the complete recognized text. The computation involves the division of the count of accurately identified words by the overall count of words present in the reference text.
\end{itemize}
The ground truth text for each of the 10 images was manually annotated and compared with the recognized text generated by the OCR model to calculate the metrics.

\section{Results and Evaluation}
The text recognition performance of the suggested model in Bengali and Nepali languages was encouraging. During training, the model achieved low Word Error Rates (WER) of 0.10 and 0.14 for Bengali and Nepali, respectively, and Character Error Rates (CER) of 0.04 and 0.10 for Bengali and Nepali, respectively. The Levenshtein distance between the recognized and ground truth texts was used to determine the CER and WER values.

\begin{figure}[h!]
\includegraphics[width=0.5\textwidth]{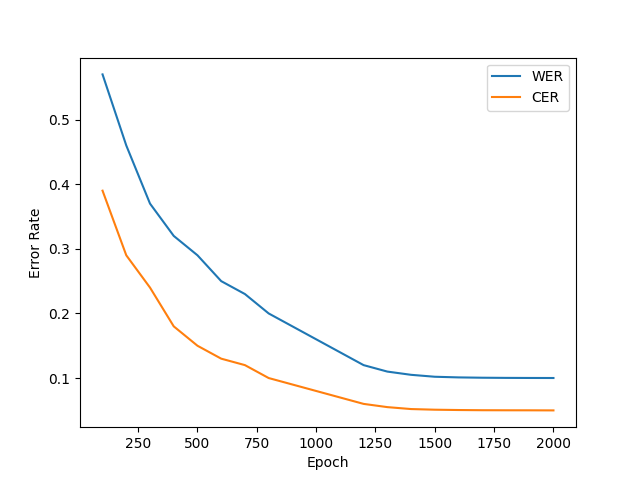}
\caption{CER and WER for Bengali on Training Dataset}
\label{fig}
\end{figure}

\begin{figure}[h!]
\includegraphics[width=0.5\textwidth]{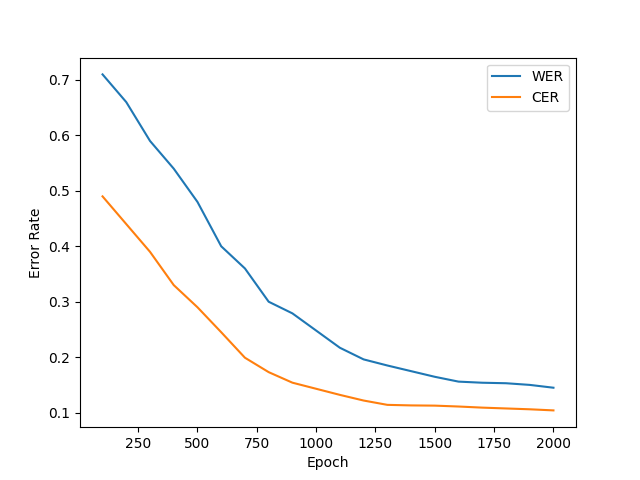}
\caption{CER and WER for Nepali on Training Dataset}
\label{fig}
\end{figure}

The CER and WER values were plotted over 2000 epochs to show the model's performance. The data demonstrated that the model, for both Bengali and Nepali languages, reached a consistently low CER and WER after roughly 1500 epochs.

The CER and WER values were marginally higher than those attained during training, though, when the model was tested on the test set. Bengali had an average CER and WER of 0.07 and 0.12, while Nepali had an average CER and WER of 0.11 and 0.15.

The modest variation in the CER and WER values between training and testing may be explained by the test set's greater diversity and changing image quality when compared to the training set. The test set may have included images with more intricate backgrounds or uncommon font styles that were not present in the training set, despite the fact that the model was trained on a varied selection of images with different font styles, sizes, and orientations.

\section{Conclusion}
In this research, a model for Bengali and Nepali text recognition was proposed. Images containing handwritten and printed text in both languages served as training data for the model. According to the experimental findings, the proposed model exhibited a low Character Error Rate (CER) and Word Error Rate (WER) during training and performed effectively on the test set.
During training, the model achieved CERs of 0.04 and 0.09, as well as WERs of 0.10 and 0.14, for Bengali and Nepali, respectively. When evaluated on the test set, the model's average CER and WER for Bengali and Nepali were 0.07 and 0.12, and 0.11 and 0.15, respectively.
These results demonstrate the model's strong performance in accurately recognizing text in Bengali and Nepali. Due to its low CER and WER values during training and its effectiveness on the test set, the model holds promise for practical applications such as document digitization and text extraction.
The suggested OCR methodology underscores its potential for practical uses like text extraction and document digitization. The model's capability to accurately identify text from diverse images in Bengali and Nepali languages illustrates its competence in handling complex and varied input.

\end{document}